# PROSTATE SEGMENTATION FROM 3D MRI USING A TWO-STAGE MODEL AND VARIABLE-INPUT BASED UNCERTAINTY MEASURE

*Huitong Pan[1], Yushan Feng[1], Quan Chen[2], Craig Meyer[1,3], and Xue Feng[1,3]*

[1]Springbok, Inc., Charlottesville, VA, USA
[2]Radiation Medicine, University of Kentucky, Lexington, KY, USA
[3]Biomedical Engineering, University of Virginia, Charlottesville, VA, USA

## ABSTRACT

This paper proposes a two-stage segmentation model, variable-input based uncertainty measures and an uncertainty-guided post-processing method for prostate segmentation on 3D magnetic resonance images (MRI). The two-stage model was based on 3D dilated U-Nets with the first stage to localize the prostate and the second stage to obtain an accurate segmentation from cropped images. For data augmentation, we proposed the variable-input method which crops the region of interest with additional random variations. Similar to other deep learning models, the proposed model also faced the challenge of suboptimal performance in certain testing cases due to varied training and testing image characteristics. Therefore, it is valuable to evaluate the confidence and performance of the network using uncertainty measures, which are often calculated from the probability maps or their standard deviations with multiple model outputs for the same testing case. However, few studies have quantitatively compared different methods of uncertainty calculation. Furthermore, unlike the commonly used Bayesian dropout during testing, we developed uncertainty measures based on the variable input images at the second stage and evaluated its performance by calculating the correlation with ground-truth-based performance metrics, such as Dice score. For performance estimation, we predicted Dice scores and Hausdorff distance with the most correlated uncertainty measure. For post-processing, we performed Gaussian filter on the underperformed slices to improve segmentation quality. Using PROMISE-12 data, we demonstrated the robustness of the two-stage model and showed high correlation of the proposed variable-input based uncertainty measures with GT-based performance. The uncertainty-guided post-processing method significantly improved label smoothness.

## 1. INTRODUCTION

For diagnosis and management of prostate diseases, prostate gland evaluation using MRI is important in clinical practice, in which prostate segmentation from 3D MRI is a critical step in quantitative analysis. As manual delineation of the prostate boundaries is time consuming and subject to inter- and intra-observer variability [1], many researches have been done using deep learning methods, such as ensemble deep convolutional neural networks [2] and convolutional neural networks with statistical shape models [3], to improve the accuracy and robustness of automatic prostate segmentation.

For prostate segmentation, as the 3D input images are often too large to fit into the memory of a typical GPU, patch-based methods by extracting and training on smaller patches can be used. However, as prostate often occupy a small region with relatively fixed location to the entire volume, such method is inefficient and suffers from limited receptive field, which is often determined by the patch size. Instead, we propose a two-stage segmentation model with the first-stage to localize the prostate using resampled low-resolution images and the second-stage to generate fine label maps using high-resolution cropped images. In the testing phase of the second stage, based on the estimated location of the prostate, cropping with randomly added variations can allow the model to 'see' the prostate in slightly different resolutions and image positions, so that the accuracy of the second stage can be improved.

Similar to other existing deep learning models, the auto-generated labels may have sub-optimal performance on certain testing cases, especially when the testing image has very different characteristics from training images, such as unclear tissue boundaries and poor image quality. As a way of estimating performance in the absence of ground truth (GT), uncertainty measures have been widely investigated [4, 5, 6], which are often calculated from the probability maps or their standard deviations from multiple model outputs for the same data. The Bayesian dropout method is often used to generate multiple testing outputs. In this study, we compared the uncertainty measures using outputs generated from the Bayesian dropout method and the proposed variable-input method. The correlations between the estimated uncertainty measures and the GT-based performance scores were used as the evaluation criteria. Furthermore, we used the most correlated measure to predict Dice score and Hausdorff distance for performance estimation. Lastly, we used Gaussian filter to auto-correct the uncertain slices which were identified by the most correlated uncertainty measure.

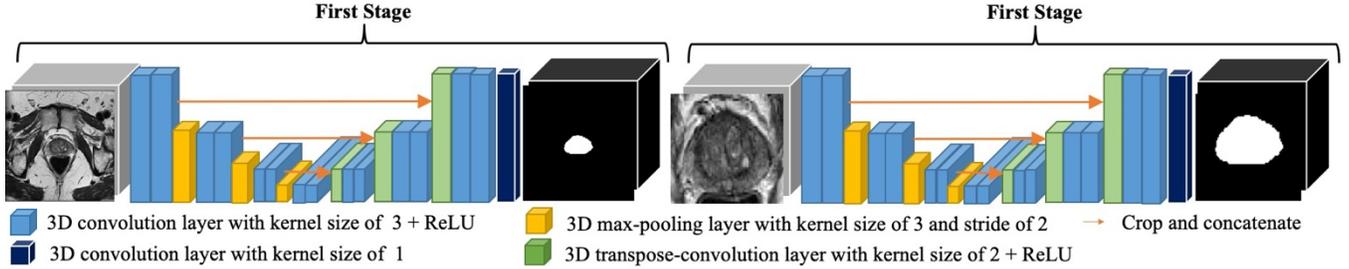

**Figure 1**: The proposed two-stage model

## 2. METHODS

We evaluated our methods on the PROMISE-12 challenge data which are T2-weighted axial MRIs acquired using various protocols [1]. Training data contains 50 cases with GT labels, and we use 40 for training and 10 for validation during the training phase. In the testing phase, we used all 50 training data to train a final model and submitted the predictions of the 30 MRIs in testing data to PROMISE-12 challenge for evaluation. For pre-processing, all images were normalized to an image spacing of 3.6mm × 0.625mm × 0.625mm.

### 2.1. Two-stage model

A two-stage model was used for prostate segmentation from 3D MRI, which included two 3D dilated U-Nets, as shown in Figure 1. The U-Nets had an encoding path on the left and a symmetric decoding path on the right. As global 3D information is important for accurate localization and segmentation of the prostate, we modified traditional 3D U-Net by replacing conventional convolutions on the encoding path with dilated convolutions to increase the receptive field without increasing the number of parameters, which was shown to be beneficial in this application. As the dimension along the slice direction is already smaller than the receptive field along this direction with the conventional convolution, the dilation factor was set to 1 × 2 × 2. The final receptive field was close to the input size.

The goal in the first stage is to get a bounding box of prostate for cropping the region of interest (ROI) in the second stage. To fit the input size into memory, we first down-sampled MRIs to 24 × 168 × 168 pixels from various original sizes, ranging from 47 × 512 × 512 to 15 × 320 × 320 pixels. We then trained a 3D dilated U-Net with 48 features at the first convolution layer.

In the second stage, as the input field-of-view was reduced with cropping, the high resolution input image could be used to obtain a finer label than the first stage. Cropping was performed based on the output of the first stage. Since first stage's incorrect localization of prostate could have a direct impact on second stage's performance, the variable-input method was used as testing augmentation for improving second stage's robustness against the first stage's localization error. The variable-input method cropped each MRI multiple times with additional random margins on top of the bounding box of first stage's predicted label map. We then resized the cropped images to 16 × 120 × 120 pixels, which was close to the average dimension for cropped images. Next, we trained another 3D dilated U-Net with 72 features at the first convolution layer. The increased number of features was used to increase the model complexity for improved performance.

Both stages were trained using the Adam optimizer with a learning rate of 0.0005, batch-normalization, 50% dropout rate at the bottom layer, Dice loss, and standard data augmentation techniques (B-Spline deformable transformation, axial flip and axial rotate).

In order to study the effectiveness of variable-input method for testing augmentation, we tested it against the Bayesian dropout method, which is commonly used for improving segmentation accuracy and for Bayesian approximation of model's uncertainty [7]. For each testing case, the experiment with variable-input method obtained 20 different outputs (probability maps) by using 10 sets of random margins and axial flip. The experiment with the Bayesian dropout method obtained 20 outputs by using fixed cropping margins, axial flip, and 50% dropout rate at the bottom block. In the end, probability maps were averaged and converted to labels with an average probability threshold of 0.5. Labels generated from the second stage were then converted to their original space and scale before cropping.

For performance evaluation, Dice score (DSC), mean surface distance (MSD), Hausdorff distance (HD) and Hausdorff distance at 95[th] percentile (HD95), were calculated based on the predicted label and GT.

### 2.2. Uncertainty measures

Since clinical evaluation of prostate segmentation is usually based on axial slices, the proposed uncertainty measures are designed as slice-based. For the input to obtain uncertainty measures, we compared probability maps generated from the Bayesian dropout method and the proposed variable-input method. To obtain the slice-wise uncertainty measures base on the given probability maps, three calculation methods were proposed, denoted as type 1 to 3 and given as follows.

**Type-1**: Mean probability of all foreground voxels. For each foreground (FG) pixel on a test slice, a mean probability was calculated from the 20 pixel-level probability maps. In order to summarize the pixel-level information into a slice-level measure, we first summed up all foreground pixels' mean probabilities, and then divided by total number of foreground pixels (N). If the model is certain, this measure should be close to 1.

**Type-2:** Mean probability variation of all foreground voxels. For each foreground pixel on a test slice, probability variation was defined as the standard deviation of the 20 pixel-level probability maps. Similar to Type-1 measure, we then summed up all probability variation and then divided the sum by N. If the model is certain, Type-2 measure should be close to 0.

**Type-3:** Pair-wise DSC. We first applied a threshold of 0.5 to convert probability maps into predicted labels. In total, 20 probability maps have 190 unique label pairs. For each pair of labels, we treated one of them as pseudo GT and calculated pair-wise pseudo DSC. Then, we averaged all pair-wise pseudo DSC as a slice-level measure.

In order to evaluate the effectiveness of uncertainty measures, we calculated the Pearson's correlation coefficient (R) between the calculated uncertainty measures and the actual performance evaluated based on the actual GT, including DSC and HD. For the final performance estimation, we predicted DSC and HD using linear regression models with their most correlated measure due to their simplicity and robustness against overfitting.

### 2.3. Post-processing

Since testing performance may vary greatly among slices, especially for apex and base slices, applying post-processing on all slices is not optimal as the performance of well-segmented slices can be negatively affected. Thus, we only post-processed the uncertain slices. Specifically, we aimed to improve label smoothness for slices with high uncertainty.

To determine the threshold for applying the post-processing, as label smoothness is mostly reflected in distance measures, slices with predicted HD above 8 mm were identified as the target slices to apply the Gaussian filter on.

### 3. EXPERIMENTS AND RESULTS

#### 3.1. Two-stage model

The first evaluation was based on the validation set containing 10 cases and results are shown in Table 1. As shown in the table, the second stage model significantly improved first stage's performance in terms of distance and DSC measures. Also, the proposed variable-input method is more effective than the dropout method for testing augmentation.

**Table 1**: Validation performance of the two-stage model

| Model | DSC | MSD | HD | HD95 |
|---|---|---|---|---|
| Stage 1 | 0.849 | 0.665 | 11.824 | 2.191 |
| Stage 2 (Variable-Input) | 0.886 | 0.449 | 7.524 | 1.435 |
| Stage 2 (Dropout) | 0.88 | 0.484 | 8.663 | 1.476 |

**Table 2**: Testing Performance in PROMISE 12

| Evaluation Metrics | Top-Ranking | 2-Stage Model | Post-Processed |
|---|---|---|---|
| Whole Prostate HD | 4.155 | 4.467 | 4.48 |
| Apex HD | 3.789 | 4.265 | 4.224 |
| Base HD | 4.41 | 4.597 | 4.74 |
| Whole Prostate DSC | 0.913 | 0.905 | 0.904 |
| Apex DSC | 0.879 | 0.861 | 0.861 |
| Base DSC | 0.897 | 0.89 | 0.889 |

**Table 3**: Correlation coefficients (R) between each uncertainty measure and actual performance scores. All p-values < 0.05.

| | | Variable-Input | | Dropout | |
|---|---|---|---|---|---|
| Uncertainty Measure | | R with DSC | R with HD | R with DSC | R with HD |
| 1 | Mean Prob. | 0.833 | -0.542 | 0.815 | -0.446 |
| 2 | Prob. Variation | -0.783 | 0.499 | -0.679 | 0.241 |
| 3 | Pair-wise DSC | 0.815 | -0.431 | 0.681 | 0.037 |

The second evaluation was based on the testing data from PROMISE12 challenge, and results are shown in Table 2. For comparison, Table 2 also reports the results from the top-ranking model as of December 2018, in the PROMISE12 challenge. The top-ranking model is an interactive segmentation method, which uses clicks along with image data as input and allows deep-learning based corrections. As shown in Table 2, with no user interaction, the proposed two-stage model only slightly underperformed the top-ranking model and still provided good performance.

#### 3.2. Uncertainty Measures

Table 3 reports the correlation (R) between uncertainty measures and slice-level GT-based performance scores for all 10 validation cases. For measure input, variable-input method had a higher correlation with the actual performance than the dropout method. In terms of calculation method, Type-1 mean probability of all foreground voxels had the highest correlation with both DSC and HD compared with mean probability variation and pair-wise DSC.

For actual performance estimation, we performed 10-fold cross validations with the 10 validation cases with 140 slices. For DSC prediction, the fitted linear regression with Type-1 measure, mean probability of all foreground pixels, had an average root mean square error (RMSE) of 0.16. For HD

**Figure 2**: Sample slices from validation data and testing data. Green contour is the ground truth label, and is not available for testing data. Blue dashed contour is the predicted label. Yellow (the lightest color) is the label contour after post-processing. For the displayed validation slice with predicted HD of 4.89mm, actual slice-level DSC improves from 0.766 to 0.816 with post-processing.

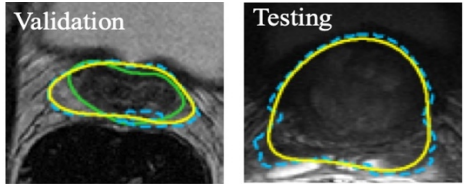

prediction, the fitted linear regression with Type-1 measure had an average RMSE of 3.85mm.

### 3.3. Targeted post-processing

In order to examine the impact of post-processing against GT, the validation set with 10 cases were split into 2 folds: 9 cases with 124 foreground slices and 1 case with 16 foreground slices. The first fold was used for fitting the linear regression, and the second fold was used for validating the post-processing method against GT.

In the second fold, the fitted linear regression identified 7 slices with predicted HD above 8 mm. The 7 identified slices were all the actual underperforming slices which had actual HD above 6 mm and actual DSC below 0.8.

Gaussian filter with sigma equal to 5 was then performed on the underperforming slices. In the second fold, targeted post-processing improved the volumetric DSC of the case from 0.873 to 0.878. Although the DSC was not significantly improved, the visual appearance of the contour was much better and clinically acceptable. One example of the post-processed slices is illustrated in Graph 2.

For testing, we fitted the linear regression models with all 10 validation cases and used them to identify underperforming slices in the final prediction on testing data, to perform post-processing. The post-processed performance scores are listed in Table 2. All the performance evaluation metrics except for apex DSC and apex HD were slightly worse. However, most of the post-processed slices had smoothness improvement and are more visually satisfactory, as illustrated in Graph 2.

### 4. DISCUSSIONS AND CONCLUSION

In this paper, we proposed an automated prostate segmentation model, uncertainty measures and uncertainty-guided post-processing methods. In terms of segmentation performance, the two-stage model basing on 3D dilated U-Net and the novel variable-input data augmentation method performed well in both validation and testing data. For uncertainty measures, the variable-input based measures outperformed the Bayesian dropout based measures, because the former resulted in higher correlation with actual GT-based performances. In terms of uncertainty metrics, the mean probability of foreground pixels was proven to be the most correlated measure with the actual performance, which is reasonable as the label foreground probability can be a good proxy for model's certainty. The predicted DSC and HD provided good estimate of the actual performance. Even though the predicted performance scores were not close to the exact numbers, the predicted scores were sufficient for differentiating well-performing and under-performing slices. For post-processing, Gaussian filter significantly improved label smoothness on uncertain slices and can be used to reduce the workload for human corrections.

In conclusion, the proposed two-stage model is a good model choice for organ segmentation. The variable-input based uncertainty measures and the related post-processing method can aid human operators when interacting with deep learning-based segmentation pipelines. All the proposed methods can also be extended to improve other applications of medical image segmentation.

### 5. REFERENCES

[1] G. Litjens, R. Toth, W. v. d. Ven, C. Hoeks and S. Kerkstra, "Evaluation of prostate segmentation algorithms for MRI: The PROMISE12 challenge," *Medical Image Analysis,* vol. 18, no. 2, pp. 359-373, 2014.

[2] H. Jia, Y. Xia, W. Cai, M. Fulham and D. D. Feng, "Prostate segmentation in MR images using ensemble deep convolutional neural networks," in *IEEE 14th International Symposium on Biomedical Imaging*, 2017.

[3] D. Karimi, G. Samei, C. Kesch, G. Nir and S. E. Salcudean, "Prostate segmentation in MRI using a convolutional neural network architecture and training strategy based on statistical shape models," *International Journal of Computer Assisted Radiology and Surgery,* vol. 13, no. 8, p. 1211–1219, 2018.

[4] B. Lakshminarayanan, A. Pritzel and C. Blundell, "Simple and Scalable Predictive Uncertainty Estimation using Deep Ensembles," in *ARXIV*, 2016.

[5] T. DeVries and G. W. Taylor, "Leveraging Uncertainty Estimates for Predicting Segmentation Quality," in *arXiv*, 2018.

[6] D. Hendrycks and K. Gimpel, "A Baseline for Detecting Misclassified and Out-of-Distribution Examples in Neural Networks," *ArXiv,* 2016.

[7] Y. Gal and Z. Ghahramani, "Dropout as a bayesian approximation: Representing model uncertainty in deep learning.," *ArXiv,* 2015.